\documentclass{article}
\usepackage[LGR,T1]{fontenc}
\usepackage[utf8]{inputenc}
\usepackage{float}
\usepackage{booktabs}
\usepackage{calc}
\usepackage{units}
\usepackage{amsmath}
\usepackage{graphicx}
\usepackage{microtype}
\usepackage[unicode=true,
 bookmarks=false,
 breaklinks=false,pdfborder={0 0 1},backref=section,colorlinks=false]
 {hyperref}

\makeatletter

\providecommand{\tabularnewline}{\\}
\floatstyle{ruled}
\newfloat{algorithm}{tbp}{loa}
\providecommand{\algorithmname}{Algorithm}
\floatname{algorithm}{\protect\algorithmname}

\PassOptionsToPackage{numbers, compress}{natbib}

     \usepackage[nonatbib,final]{neurips_bdl2019}

\usepackage{url}
\usepackage{amsfonts}
\usepackage{nicefrac}

\title{Stochastic Variational Inference via Upper Bound}

\author{%
  Chunlin ~Ji\\
  Kuang-Chi Institute of Advanced Technology\\
  Shenzhen, China\\
  \texttt{chunlin.ji@kuang-chi.org} \\
   \And
  Haige ~Shen\\
  ZenRhyme Consulting Services \\
  Shanghai, China \\
  \texttt{haige.shen@zenrhyme.com} \\
}

\makeatother

\begin{document}
\maketitle
\begin{abstract}
Stochastic variational inference (SVI) plays a key role in Bayesian
deep learning. Recently various divergences have been proposed to
design the surrogate loss for variational inference. We present a
simple upper bound of the evidence as the surrogate loss. This evidence
upper bound (EUBO) equals to the log marginal likelihood plus the
KL-divergence between the posterior and the proposal. We show that
the proposed EUBO is tighter than previous upper bounds introduced
by $\chi$-divergence or $\alpha$-divergence. To facilitate scalable
inference, we present the numerical approximation of the gradient
of the EUBO and apply the SGD algorithm to optimize the variational
parameters iteratively. Simulation study with Bayesian logistic regression
shows that the upper and lower bounds well sandwich the evidence and
the proposed upper bound is favorably tight. For Bayesian neural network,
the proposed EUBO-VI algorithm outperforms state-of-the-art results for various
examples.
\end{abstract}

\section{Introduction}

Stochastic variational inference (SVI) plays a key role in Bayesian
deep learning. SVI solves the Bayesian inference problem by introducing
a variational distribution $q(\theta;\lambda)$ over the latent variables
$\theta$ \cite{Jordan_99,Hoffman_etal_2013}, and then minimizes
the Kullback-Leibler (KL) divergence between the approximating distribution
$q(\theta;\lambda)$ and the exact posterior $p(\theta|\mathcal{D})$.
This minimization is the same as maximizing the evidence lower bound
(ELBO), $\mathcal{L}=\int q(\theta;\lambda)[\log p(\mathcal{D},\theta)-\log q(\theta;\lambda)$
\cite{Jordan_99}, which is a lower bound of the model evidence $\log p(\mathcal{D})$.
The KL-divergence is the discrepancy between the surrogate loss ELBO
and the evidence. SVI turns the Bayesian inference into an optimization
problem. It has been recognized that the property of discrepancy has
strong influence on the optimization of ELBO \cite{Li_Turner_2017,Dieng_etal_2017,Zhang_etal_2018}.
The classical objective $D_{\mathrm{KL}}(q(\theta;\lambda)||p(\theta|\mathcal{D}))$
leads to approximate the posteriors with zero-forcing behavior. This
zero-forcing behavior imposes undesirable properties, which may lead
to underestimation of the posterior’s support especially when dealing
with light-tailed posteriors or multi-modal posteriors \cite{Murphy_2012,Hensman_etal_2014,Dieng_etal_2017}.
Various SVI methods are proposed to use different divergences to achieve
tighter bound and/or mass-covering property, which may lead to better
performance in SVI \cite{Li_Turner_2017,Dieng_etal_2017,Tao_etal_2018_VI}.
Recently, the upper bound of $\log p(D)$ has captured certain attention.
For instance, the variational Rényi bound ($\mathcal{L}_{\alpha}(\lambda)=\frac{1}{1-\alpha}\log E_{q}[(\frac{p(\mathcal{D},\theta)}{q(\theta;\lambda)})^{1-\alpha}]$,
for $\alpha\neq1$) and the $\chi$ upper bound ($\mathcal{L}_{\chi^{n}}(\lambda)=\frac{1}{n}\log E_{q}[(\frac{p(\mathcal{D},\theta)}{q(\theta;\lambda)})^{n}]$,
for $n>1$) are introduced to provide the mass-covering property:
optimizing such divergence leads to a variational distribution with
a mass-covering (or say zero-avoiding) behavior. The resulting gradient
of $\alpha$-divergence based upper bound is,
\begin{equation}
\mathcal{\nabla_{\lambda}L}_{\alpha}=\frac{1}{1-\alpha}\mathbb{E}_{q}\left[w_{\alpha}\nabla_{\lambda}(\log p(\mathcal{D},\theta)-\log q(\theta;\lambda))\right]\label{eq:ELBO}
\end{equation}
where $\nicefrac{w_{\alpha}=\left(\frac{p(\mathcal{D},\theta)}{q(\theta;\lambda)}\right)^{1-\alpha}}{E_{q}\left[\left(\frac{p(\mathcal{D},\theta)}{q(\theta;\lambda)}\right)^{1-\alpha}\right]}$
\cite{Li_Turner_2017}. Interestingly, we have the connection that
when $\alpha\rightarrow1$, the Rényi- divergence become the KL-divergence
and the Rényi-VI algorithm reduces to the standard SVI \cite{Hoffman_etal_2013};
when $\alpha=0$ , the Rényi-VI becomes the importance weighted auto-encoder
(IWAE) or say the importance weighted VI\cite{Burda_etal_2015IWVAE,Domke_2018_IWVI};
when $\alpha=-1$ , it becomes the $\chi^{2}$ divergence based CHIVI
algorithm \cite{Dieng_etal_2017}. The integration in the gradient
$\mathcal{\nabla_{\lambda}L}_{\alpha}$ can be estimated by Monte
Carlo approach with samples drawn from $q(\theta;\lambda)$. Then
the stochastic gradient descendant (SGD) algorithm is applied to optimize
the variational parameter $\lambda$. Moreover, to deal with large
datasets, we may use a minibatch of the data $\mathcal{D}$ in the
evaluation of the joint distribution $p(\mathcal{D},\theta)$ \cite{Ranganath_etal_2014}.
Inspired by these previous works, here we explore a simple but effective
upper bound for variational inference, our main contributions are
summarized:

1. We propose a new upper bound of the evidence by introducing the
KL-divergence between the posterior and the variational distribution.
The proposed EUBO possess the mass covering properties, and is tighter
then the upper bounds introduced by $\alpha$-divergence or $\chi$-divergence.

2. We present the numerical approximation of the gradient of the EUBO,
apply the SGD algorithm to optimize the variational parameters, and
format a black-box style variational inference which is suitable for
scalable Bayesian inference.

3. With the upper and lower bound sandwiching the evidence, we can
be more confident to evaluate the fitting of model. We also find that
the EUBO converges faster and is tighter than the ELBO. In Bayesian
neural network regression, the EUBO-VI algorithm gains improvement
in both the test error and the model fitting.

\section{Variational Inference with the EUBO}

\subsection{Upper bound of the model evidence}

As it is well known that the KL-divergence is not symmetric, $D_{\mathrm{KL}}(q||p)$
possesses the zero-force property, but the inverse KL-divergence $D_{\mathrm{KL}}(p||q)$
posses the mass-covering property. This motivates our work to leverage
on $D_{\mathrm{KL}}(p||q)$ in the design of the surrogate loss. By
using Gibbs' inequality that $-\int p(\theta|\mathcal{D})\log p(\theta|\mathcal{D})d\theta\leq-\int p(\theta|\mathcal{D})\log q_{\lambda}(\theta)d\theta$
holds for any distribution $q(\theta;\lambda)$, we introduce an upper
bound on the model evidence,
\begin{align}
\log p(\mathcal{D}) & =\int p(\theta|\mathcal{D})\log p(\mathcal{D},\theta)d\theta-\int p(\theta|\mathcal{D})\log p(\theta|\mathcal{D})d\theta\nonumber \\
 & \leq\int p(\theta|\mathcal{D})\log p(\mathcal{D},\theta)d\theta-\int p(\theta|\mathcal{D})\log q(\theta;\lambda)d\theta=\int p(\theta|\mathcal{D})\log\frac{p(\mathcal{D},\theta)}{q(\theta;\lambda)}
\end{align}
We define $\mathcal{U}(\lambda)\equiv\int p(\theta|\mathcal{D})\log\dfrac{p(\mathcal{D},\theta)}{q(\theta;\lambda)}d\theta$
as the Evidence Upper BOund (EUBO). It is easy to find that $\mathcal{U}(\lambda)=\log p(\mathcal{D})+\mathrm{D_{KL}}(p(\theta|\mathcal{D})||q(\theta;\lambda))$,
that is $D_{\mathrm{{KL}}}(p(\theta|\mathcal{D})||q(\theta;\lambda))$
just the discrepancy between the EUBO $\mathcal{U}(\lambda)$ and
the true $\log p(\mathcal{D})$.

\subsection{Some properties of the EUBO}

We present three favorable properties of the proposed EUBO $\mathcal{U}(\lambda)$.
First the discrepancy $D_{\mathrm{{KL}}}(p(\theta|\mathcal{D})||q(\theta;\lambda))$
possesses mass covering property. According to previous studies\cite{Li_Turner_2017,Dieng_etal_2017},
the mass covering property has advance in approximation of the posterior.
The mass covering property of EUBO, can be easily verified by simulation
examples \cite{Minka_05_Divergence,Ji_06_adaptiveMC}. Second, $D_{\mathrm{{KL}}}(p(\theta|\mathcal{D})||q(\theta;\lambda))$
is a tight discrepancy. We find that the $D_{\mathrm{{KL}}}(p(\theta|\mathcal{D})||q(\theta;\lambda))$
proposed in this work is tighter than the $\alpha$-divergence and
$\chi^{n}$-divergence. With this tighter discrepancy, we can then
design tighter upper bounds. As discussed in \cite{Burda_etal_2015IWVAE,Tao_etal_2018_VI},
tighter bound tends to lead better performance in SVI. We present
a theorem to show the advantage of the proposed EUBO (see the appendix
for a brief proof).

\textbf{Theorem 1}. Define the EUBO as equation (2). Then the following
holds:
\begin{itemize}
\item ELBO $\mathcal{L}(\lambda)$ $\leq\log p(\mathcal{D})\leq$ EUBO $\mathcal{U}(\lambda)$.
\item $D_{\alpha}(q(\theta)||p(\theta|\mathcal{D}))\geq\frac{-\alpha}{1-\alpha}D_{\mathrm{{KL}}}(p(\theta|\mathcal{D})||q(\theta))$
, for $\alpha\leq0$; $D_{\chi^{n}}(q(\theta)||p(\theta|\mathcal{D}))\geq\frac{n-1}{n}D_{\mathrm{{KL}}}(p(\theta|\mathcal{D})||q(\theta))$,
for $n\geq2$.
\item $\frac{-\alpha}{1-\alpha}\mathcal{U}(\lambda)+\frac{1}{1-\alpha}\log p(\mathcal{D})\leq\mathcal{U}_{\alpha}(\lambda)$,
for $\alpha\leq0$ ; $\frac{n-1}{n}\mathcal{U}(\lambda)+\frac{1}{n}\log p(\mathcal{D})\leq\mathcal{U}_{\chi^{n}}(\lambda)$,
for $n\geq2$.
\end{itemize}
Third, the numerical estimation of EUBO is close to the ground true
evidence, even when the variational distribution has not been well
optimized. The EUBO involves an integration with respect to the unknown
posterior. As stated in later section, we use importance sampling
in the estimation of this upper bound. The using of importance sampling
reduces the bias of the MC estimation of the evidence when the variational
proposal has certain discrepancy with the posterior. This property
leads to two advantage: 1) model selection based solely on the ELBO
is inappropriate because of the possible variation in the tightness
of this bound, with the accompanying upper bound to sandwich the evidence,
one can perform model selection with more confidence\cite{Grosse_etal_2015,Ji_2010_Bounded};
2) in case we need to evaluate the evidence or its surrogate when
the optimization process of variational parameters can run only few
steps, for example in the Bayesian Meta-learning\cite{Kim_2018_BayesianMAML,Ravi_2019_AmortizedBMAML},
the proposed EUBO may be preferable since its estimation is insensitive
to the bias of variational distribution.

\subsection{SGD for the EUBO}

By definition, the EUBO $\mathcal{U}(\lambda)$ has an integration
with respect to unknown posterior. In previous study\cite{Ji_2009_Thesis,Ji_2010_Bounded},
the posterior is represented by simulated sample from Markov chain
Monte Carlo (MCMC), and the integration is approximated via Monte
Carlo. However, Monte Carlo particularly MCMC is not a scalable method
in dealing with large scale Bayesian learning problem. Following the
idea of black-box VI\cite{Ranganath_etal_2014}, we use only a few
number of Monte Carlo samples to obtain a `noisy' gradient of the
EUBO, and then apply the SGD approach to optimize $\lambda$ iteratively.
First, we derive the gradient of $\mathcal{U}(\lambda)$ (refer to
the Appendix for details), denoted by $\nabla_{\lambda}\mathcal{U}$,
\begin{equation}
\nabla_{\lambda}\mathcal{U}=\mathbb{E}_{q}\left\{ w(\theta)\left[(\log\omega(\theta)+1)\nabla_{\lambda}\log p(\mathcal{D},\theta)-\nabla_{\lambda}\log q(\theta;\lambda)\right]\right\} \label{eq:gradient_EUBO}
\end{equation}
where $w(\theta)=\frac{p(\theta|D)}{q(\theta;\lambda)}$ and $\omega(\theta)=\frac{p(D,\theta)}{q(\theta;\lambda)}.$
Note that the posterior $p(\theta|\mathcal{D})$ in $w(\theta)$ is
generally unknown, so we use the joint distribution $p(\mathcal{D}|\theta)p(\theta)$
instead and normalizes the weights to cancel the unknown constant
$p(\mathcal{D})$. Given the samples $\{\theta^{(i)}\}_{i=1}^{M}$
drawn from $q(\theta;\lambda)$ and normalized weights $\{\hat{w}^{(i)}\}_{i=1}^{M}$,
we estimate $\nabla_{\lambda}\mathcal{U}$ as follows,
\begin{equation}
\hat{\nabla}_{\lambda}\mathcal{U}=\sum_{i}^{M}\hat{w}^{(i)}\left[(\log\omega(\theta^{(i)})+1)\nabla_{\lambda}\log p(\mathcal{D},\theta^{(i)})-\nabla_{\lambda}\log q(\theta^{(i)};\lambda)\right].\label{eq:est_grad_EUBO}
\end{equation}
To deal with large dataset, we divide the entire dataset to mini-batch.
Only a mini-batch data $\mathcal{S}=\{x_{n}\}_{n=1}^{S}$ is used
in the evaluation of $p(\mathcal{D},\theta)$,
\[
p(\mathcal{D},\theta^{(i)})\approx\left[\prod_{n=1}^{S}p(x_{n}|\theta^{(i)})\right]^{\frac{N}{S}}p(\theta^{(i)}).
\]

To incorporate with the advantage of autogradient packages \cite{},
we take the reparametrization trick \cite{Kingma_2014_VAE} that $\theta^{(i)}=g_{\lambda}(\epsilon^{(i)})=\mu+\sigma\text{\ensuremath{\epsilon^{(i)}}}$,
where $\lambda=[\mu,\sigma]$ and $\epsilon^{(i)}\sim\mathcal{N}(0,1)$,
then the resulting reparameterization gradient becomes,
\begin{equation}
\hat{\nabla}_{\lambda}\mathcal{U}=\sum_{i}^{M}\hat{w}^{(i)}\left[(\log\omega(\theta^{(i)})+1)\nabla_{\lambda}\log p(\mathcal{D},g_{\lambda}(\epsilon^{(i)}))-\nabla_{\lambda}\log q(g_{\lambda}(\epsilon^{(i)});\lambda)\right].\label{eq:est_grad_EUBO-1}
\end{equation}
This reparametrization trick significantly reduces the effort needed
to implement variational inference in a wide variety of models. Finally,
with the noisy estimation of the gradient $\hat{\nabla}_{\lambda}\mathcal{U}$,
the SGD algorithm is applied to minimize the EUBO iteratively, $\lambda_{t+1}=\lambda_{t}-\alpha*\hat{\nabla}_{\lambda}\mathcal{U}$.
The algorithm is presented as follows,

\begin{algorithm}
\caption{EUBO based variational inference (EUBO-VI) }

\begin{itemize}
\item Initialization: initialize the parameter $\lambda_{0}$ randomly or
by some pre-specified values, the learning rate $\alpha_{0}$.
\item for $t=0,...,T$
\begin{itemize}
\item pick up a mini-batch data $\mathcal{S}=\{x_{n}\}_{n=1}^{S}$.
\item generate $M$ samples $\{\theta^{(i)}\}_{i=1}^{M}$ : $\epsilon^{(i)}\sim\mathcal{N}(0,1)$
, $\theta^{(i)}=g_{\lambda}(\epsilon^{(i)}=\mu+\sigma\text{\ensuremath{\epsilon^{(i)}}}$.
\item calculate the weight $\log w^{\ensuremath{(i)}}=\frac{N}{S}\sum_{n=1}^{S}\log p(x_{n}|\theta^{(i)})+\log p(\theta^{(i)})-\log q(\theta^{(i)};\lambda_{t})$,
$w^{\ensuremath{(i)}}=\exp\left(\log w^{(i)}-\max\{\log w^{(i)}\}\right)$,
and normalize $\hat{w}^{(i)}=\frac{w^{(i)}}{\sum_{i}w^{(i)}}$ .
\item evaluate the gradient $\hat{\nabla}_{\lambda}\mathcal{U}=-\sum\hat{w}^{(i)}\nabla_{\lambda}\log q(g_{\lambda}(\epsilon^{(i)})$
.
\item update $\lambda_{t+1}=\lambda_{t}-\alpha_{t}*\hat{\nabla}_{\lambda}\mathcal{U}$
.
\end{itemize}
\end{itemize}
\end{algorithm}

After we obtain the optimal $\lambda^{*}$, the resulting minimum
upper bound $\mathcal{U}(\lambda)$ is then estimated by
\begin{equation}
\mathcal{\hat{U}^{*}}=\sum_{i=1}^{M}\hat{w}^{(i)}\left[\sum_{n=1}^{N}\log p(x_{n}|\theta^{(i)})+\log p(\theta^{(i)})-\log q(\theta^{(i)};\lambda^{*})\right]\label{eqs:upperbound}
\end{equation}

Furthermore, if we assume the posterior $p(\theta|\mathcal{D})$ and
the joint distribution $p(\theta,\mathcal{D})$ has no relation to
the variational parameter $\lambda$ , which means $\nabla_{\lambda}p(\theta|\mathcal{D})=0$
and $\nabla_{\lambda}\log p(\mathcal{D},\theta)=0$, then we obtain
the score gradient, $\hat{\nabla}_{\lambda}\mathcal{U}=-\sum_{i}^{M}\hat{w}^{(i)}\nabla_{\lambda}\log q(\theta^{(i)};\lambda)$.
Compare with the score gradient of CHIVI\cite{Dieng_etal_2017}, we
find the only difference is the definition of $\hat{w}$: $\hat{w}\propto\frac{p(D,\theta)}{q(\theta;\lambda)}$
for our EUBO-VI, while $\hat{w}\propto(\frac{p(D,\theta)}{q(\theta;\lambda)})^{2}$
for CHIVI. However, although there is some similarity between the
EUBO-VI and CHIVI/Rényi-VI, EUBO-VI is unique and does not fall into
any special case of these previous upper bound based SVI algorithms.

\section{Simulation studies}

\subsection{Bayesian Logistic Regression}

For Bayesian logistic regression, we use data sets from the UCI repository:
Iris, Pima, Spectf, Wdbc and Ionos. The variate dimension in these
data sets range from 5 to 45, including a dimension of all ones to
account for offset. We set the prior distribution of the coefficients
as $\mathcal{N}(0,1)$, choose the number of importance sampling samples
as 10 and the mini-batch size as 100. In our experiments we use the
Adam algorithm \cite{Kingma_2014_Adam}, an adaptive version of SGD
which automatically tune the learning rates according to the history
of gradients and their variances. We perform experiments with the
proposed EUBO-SVI, in comparision with vanilla SVI \cite{Paisley_etal_2012},
CHIVI\cite{Dieng_etal_2017} and Rényi-VI \cite{Li_Turner_2017}.
We also compared the optimized bounds obtained by various algorithms:
$\mathcal{\hat{L}}$ denotes ELBO from vanilla SVI, $\mathcal{\hat{U}}$
denotes EUBO from the proposed algorithm, $\mathcal{U}_{\chi^{2}}$
denotes the upper bound from CHIVI, $\mathcal{\hat{U}}_{\alpha=-2}$
and $\mathcal{\hat{L}}_{\alpha=2}$ denotes the upper and lower bounds
from the Rényi-VI with different $\alpha$-divergences.

Take the logistic regression of Iris dataset as example, we show the
estimated bound of ELBO and EUBO of each epoch in Figure 1. We observed
that the EUBO converges faster, and is tighter than the ELBO even
in the early stage when the variational parameters are not well optimized.
We run 20 trails for each bounds, the results are shown in table 1.
The upper and lower bounds well bracket the log evidence. With both
the lower and upper bounds being close to each other, we are more
confidence that the VIs find the true evidence. Moreover, simulation
results confirm that $\frac{1}{2}\mathcal{U}(\lambda)+\frac{1}{2}\log p(\mathcal{D})\leq\mathcal{U}_{\chi^{2}}(\lambda)$,
which is consistent with Theorem 1 discussed in Section 2.2. To test
the performance of prediction, all the datasets are randomly partitioned
into 90\% for training and 10\% for testing, and the results are averaged
over 20 random trials. The averaged test error is shown in table 2,
which shows that the EUBO-SVI algorithm performs well in model prediction.

\begin{figure}
\centering %
\fbox{\includegraphics[width=0.6\linewidth]{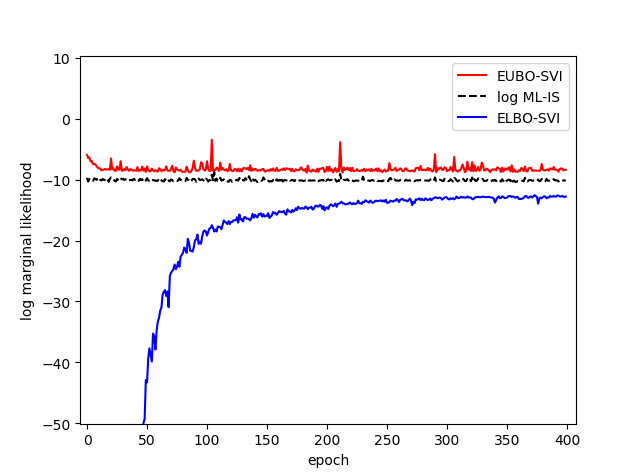}} \caption{Bayesian logistic regression on Iris data: the upper and lower bounds
sandwich the log marginal likelihood.}
\end{figure}

\begin{table}
\caption{Bayesian logistic regression: estimated upper and lower bounds}
\label{sample-table} \centering%
\begin{tabular}{lccccc}
\toprule
 & Iris  & Pima  & Spectf  & Wdbc  & Ionos\tabularnewline
\midrule
$\mathcal{\hat{U}}_{\alpha=-2}$  & -7.94$\pm$0.70  & -41$\pm$1.96  & -6.20$\pm$1.79  & -6.97$\pm$2.24  & -28.11$\pm$2.29\tabularnewline
$\mathcal{\hat{U}}$  & -8.34$\pm$0.37  & -44.7$\pm$2.79  & -7.53$\pm$1.60  & -8.38$\pm$2.39  & -30.01$\pm$2.46\tabularnewline
$\mathcal{U}_{\chi^{2}}$  & -8.58$\pm$0.71  & -45.3$\pm$2.07  & -8.33$\pm$1.05  & $-9.53\pm$1.95  & -34.67$\pm$2.11\tabularnewline
$\frac{1}{2}\mathcal{\hat{U}}+\frac{1}{2}\log\hat{p}(\mathcal{D})$  & -9.18$\pm$0.23  & -46.84$\pm$2.21  & -9.02$\pm$1.12  & -10.84$\pm$2.02  & -33.86$\pm$2.18\tabularnewline
$\log\hat{p}(\mathcal{D})$  & -10.03$\pm$0.17  & -48.94$\pm$1.65  & -10.51$\pm$0.65  & -13.3$\pm$1.65  & -37.72$\pm$1.91\tabularnewline
$\mathcal{\hat{L}}$  & -12.85$\pm$0.27  & -53.36$\pm$0.95  & -12.58$\pm$0.30  & -18.99$\pm$0.29  & -46.67$\pm$0.40\tabularnewline
$\mathcal{\hat{L}}_{\alpha=2}$  & -14.29$\pm$0.55  & -55.55$\pm$1.29  & -13.19$\pm$0.21  & -25.18$\pm$1.04  & -51.01$\pm$0.45\tabularnewline
\bottomrule
\end{tabular}
\end{table}

\begin{table}
\caption{Bayesian logistic regression: test error}
\centering%
\begin{tabular}{cccccc}
\toprule
 & Iris  & Pima  & Spectf  & Wdbc  & Ionos\tabularnewline
\midrule
$D_{\alpha=-2}$-SVI  & 0.0$\pm$0.0  & 0.240$\pm$0.043  & 0.085$\pm$0.046  & 0.019$\pm$0.013  & 0.128$\pm$0.070\tabularnewline
CHIVI  & 0.0$\pm$0.0  & 0.230$\pm$0.038  & 0.080$\pm$0.062  & \textbf{0.013$\pm$0.013}  & 0.090$\pm$0.039\tabularnewline
EUBO-SVI  & \textbf{0.0$\pm$0.0 } & 0.231$\pm$0.045  & \textbf{0.080$\pm$0.055}  & 0.014$\pm$0.012  & \textbf{0.085$\pm$0.045}\tabularnewline
ELBO-SVI  & 0.0$\pm$0.0  & 0.235$\pm$0.040  & 0.083$\pm$0.058  & 0.017$\pm$0.012  & 0.089$\pm$0.046\tabularnewline
$D_{\alpha=2}$-SVI  & 0.0$\pm$0.0  & \textbf{0.229$\pm$0.042}  & 0.081$\pm$0.057  & 0.014$\pm$0.014  & 0.088$\pm$0.0394\tabularnewline
\bottomrule
\end{tabular}
\end{table}

\subsection{Bayesian Neural Network}

In the Bayesian neural network regression, we take the same setting
with previous work \cite{Li_Turner_2017}. We use neural networks
with one hidden layers, and take 50 hidden units for most datasets,
except that we take 100 units for Protein which are relatively large;
We set $\mathcal{N}(0,1)$ as the prior distribution for the weight
and bias of the neural network. We choose $RELU(x)=max(0,x)$ as the
active function. The number of importance sampling samples is 10 and
the mini-batch size is 100. All the datasets are randomly partitioned
into 90\% for training and 10\% for testing, and the results are averaged
over 20 random trials. We compare our EUBO-VI with state-of-the-art
results directly cited from some representative algorithms: Probabilistic
backpropagation(BPB)\cite{HernndezLobato_2015_ProbabilisticBP},
Rényi-VI \cite{Li_Turner_2017}, and CLBO-VI \cite{Tao_etal_2018_VI}.
The averaged prediction accuracy and averaged test log likelihood
(LL) are given in Table 3 and Table 4 respectively. The proposed EUBO-VI
achieves substantial improvement on both the test error and negative
log likelihood for various examples.

\begin{table}
\caption{Bayesian neural network regression: average test RMSE}
\label{sample-table-1} \centering%
\begin{tabular}{lccccc}
\toprule
 & EUBO-VI  & Rényi-VI\cite{Tao_etal_2018_VI}  & CLBO-VI \cite{Tao_etal_2018_VI} & ELBO-VI\cite{Li_Turner_2017}  & BPB\cite{HernndezLobato_2015_ProbabilisticBP}\tabularnewline
\midrule
Boston  & \textbf{2.62$\pm$0.16}  & 2.86$\pm$0.40  & 2.71$\pm$0.29  & 2.89$\pm$0.17  & 2.977$\pm$0.093\tabularnewline
Concrete  & \textbf{4.81$\pm$0.22}  & 5.15$\pm$0.25  & 5.04$\pm$0.27  & 5.42$\pm$0.11  & 5.506$\pm$0.103\tabularnewline
Energy  & 0.94$\pm$0.08  & 1.00$\pm$0.18  & 0.95$\pm$0.15  & \textbf{0.51$\pm$0.01}  & 1.734$\pm$0.051\tabularnewline
Kin8nm  & \textbf{0.08$\pm$0.00}  & 0.08$\pm$0.00  & 0.08$\pm$0.00  & 0.08$\pm$0.00  & 0.098$\pm$0.001\tabularnewline
Naval  & 0.01$\pm$0.01  & 0.00$\pm$0.01  & 0.00$\pm$0.00  & \textbf{0.00$\pm$0.00}  & 0.006$\pm$0.000\tabularnewline
Combined  & 4.07$\pm$0.10  & 4.13$\pm$0.04  & \textbf{4.03$\pm$0.06 } & 4.07$\pm$0.04  & 4.052$\pm$0.031\tabularnewline
Protein  & \textbf{4.41$\pm$0.05 } & 4.65$\pm$0.07  & 4.43$\pm$0.05  & 4.45$\pm$0.02  & 4.623$\pm$0.009\tabularnewline
Wine  & \textbf{0.60$\pm$0.03}  & 0.62$\pm$0.03  & 0.61$\pm$0.03  & 0.63$\pm$0.01  & 0.614$\pm$0.008\tabularnewline
Yacht  & \textbf{0.75$\pm$0.05}  & 0.94$\pm$0.23  & 0.87$\pm$0.18  & 0.81$\pm$0.05  & 0.778$\pm$0.042\tabularnewline
\bottomrule
\end{tabular}
\end{table}

\begin{table}
\caption{Bayesian neural network regression: average negative test LL(lower
is better)}
\centering%
\begin{tabular}{cccccc}
\toprule
 & EUBO-SVI  & Rényi-VI\cite{Tao_etal_2018_VI}  & CLBO-VI\cite{Tao_etal_2018_VI}  & ELBO-VI\cite{Li_Turner_2017}  & BPB\cite{HernndezLobato_2015_ProbabilisticBP}\tabularnewline
\midrule
Boston  & \textbf{2.37$\pm$0.02}  & 2.46$\pm$0.16  & 2.40$\pm$0.09  & 2.52$\pm$0.03  & 2.579$\pm$0.052\tabularnewline
Concrete  & \textbf{2.83$\pm$0.03}  & 3.04$\pm$0.07  & 3.02$\pm$0.03  & 3.11$\pm$0.02  & 3.137$\pm$0.021\tabularnewline
Energy  & 1.63$\pm$0.03  & 1.67$\pm$0.05  & 1.65$\pm$0.04  & \textbf{0.77$\pm$0.02}  & 1.981$\pm$0.023\tabularnewline
Kin8nm  & \textbf{-1.16$\pm$0.01}  & -1.14$\pm$0.02  & -1.14$\pm$0.02  & -1.12$\pm$0.01  & -0.901$\pm$0.010\tabularnewline
Naval  & -3.81$\pm$0.05  & -4.11$\pm$0.11  & -4.17$\pm$0.01  & \textbf{-6.49$\pm$0.04}  & -3.735$\pm$0.004\tabularnewline
Combined  & 2.82$\pm$0.01  & 2.84$\pm$0.04  & \textbf{2.81$\pm$0.02 } & 2.82$\pm$0.01  & 2.819$\pm$0.008\tabularnewline
Protein  & \textbf{2.87$\pm$0.02 } & 2.93$\pm$0.00  & 2.89$\pm$0.01  & 2.91$\pm$0.00  & 2.950$\pm$0.002\tabularnewline
Wine  & \textbf{0.92$\pm$0.03} & 0.94$\pm$0.04 & 0.93$\pm$0.04  & 0.96$\pm$0.01  & 0.931$\pm$0.014\tabularnewline
Yacht  & \textbf{1.12$\pm$0.02}  & 1.61$\pm$0.00  & 1.52$\pm$0.00  & 1.77$\pm$0.01  & 1.211$\pm$0.044\tabularnewline
\bottomrule
\end{tabular}
\end{table}

\section{Conclusion}

We investigate the variational inference utilize an effective evidence
upper bound (EUBO). The proposed upper bound is based on the KL-divergence
between the posterior and variational distribution. We real that this
upper bound is tighter than the previous Rényi bound and $\chi$-bound.
We proposed the SGD algorithm to optimize the EUBO and develop the
reparametrization trick for easy implementation using prevalent python
packages for large scale problems. We compare the proposed algorithm
with vanilla SVI, CHIVI and Rényi-VI. Simulation study shows that
the proposed EUBO-VI algorithm gains improvement on both the test
error and log likelihood. We also observed that the EUBO converges
faster, and is tighter than the ELBO even in the early stage of the
optimization procedure. Moreover, with the both upper and lower bound,
we are much confidence with the model fitness. This upper bound VI
not only provided for variational inference for complex model, but
also provide an easy accessible upper bound for model criticism.

 \bibliographystyle{plain}
\bibliography{mybib}

\section*{Appendix}

\subsection*{Gradient of the EUBO}

We present two ways to derive the gradient of the EUBO $\mathcal{U}_{\lambda}$.
Reform $\mathcal{U}_{\lambda}$ as an integration with $q(\theta;\lambda)$,
that is $\int q(\theta;\lambda)\left(\frac{p(\theta|\mathcal{D})}{q(\theta;\lambda)}\right)\log\frac{p(\mathcal{D},\theta)}{q(\theta;\lambda)}d\theta$,
and denote $w(\theta)=\frac{p(\mathcal{D},\theta)}{q(\theta;\lambda)}$
, $\hat{w}(\theta)=\frac{p(\theta|\mathcal{D})}{q(\theta;\lambda)}$
. Note that $\nabla_{\lambda}\log w(\theta)=\nabla_{\lambda}\log\hat{w}(\theta)+\nabla_{\lambda}\log p(D)=\nabla_{\lambda}\log\hat{w}(\theta)$.
We can derive the gradient as follows, {\small{}{}{}{}
\begin{eqnarray*}
\nabla_{\lambda}\mathcal{U} & = & \nabla_{\lambda}\int q(\theta;\lambda)\hat{w}(\theta)\log w(\theta)d\theta\\
 & = & \int\nabla_{\lambda}\left[q(\theta;\lambda)\hat{w}(\theta)\log w(\theta)\right]d\theta\\
 & = & \int\nabla_{\lambda}q(\theta;\lambda)\hat{w}(\theta)\log w(\theta)d\theta+\int q(\theta;\lambda)\nabla_{\lambda}\hat{w}(\theta)\log w(\theta)d\theta+\int q(\theta;\lambda)\hat{w}(\theta)\nabla_{\lambda}\log w(\theta)d\theta\\
 & = & \mathbb{E}_{q}\left[\hat{w}(\theta)\log w(\theta)\nabla_{\lambda}\text{\ensuremath{\log}}q(\theta;\lambda)\right]+\mathbb{E}_{q}\left[\log w(\theta)\nabla_{\lambda}\hat{w}(\theta)\right]+\mathbb{E}_{q}\left[\hat{w}(\theta)\nabla_{\lambda}\log w(\theta)\right]\\
 & = & \mathbb{E}_{q}[\hat{w}(\theta)\log w(\theta)\nabla_{\lambda}\text{\ensuremath{\log}}q(\theta;\lambda)]+\mathbb{E}_{q}[\log w(\theta)\hat{w}(\theta)\nabla_{\lambda}\log w(\theta)]+\mathbb{E}_{q}[\hat{w}(\theta)\nabla_{\lambda}\log w(\theta)]\\
 & = & \mathbb{E}_{q}[\hat{w}(\theta)\log w(\theta)\nabla_{\lambda}\text{\ensuremath{\log}}q(\theta;\lambda)]+\mathbb{E}_{q}[\log w(\theta)\hat{w}(\theta)\text{(\ensuremath{\nabla_{\lambda}}}\log p(\mathcal{D},\theta)-\nabla_{\lambda}\log q(\theta;\lambda))]+\mathbb{E}_{q}[\hat{w}(\theta)\nabla_{\lambda}\log w(\theta)]\\
 & = & \mathbb{E}_{q}[\hat{w}(\theta)\log w(\theta)\nabla_{\lambda}\log p(\mathcal{D},\theta)\text{]}+\mathbb{E}_{q}[\hat{w}(\theta)\text{[\ensuremath{\nabla_{\lambda}}}\log p(\mathcal{D},\theta)-\nabla_{\lambda}\log q(\theta;\lambda)]d\theta\\
 & = & \mathbb{E}_{q}[\hat{w}(\theta)(\log w(\theta)+1)\nabla_{\lambda}\log p(\mathcal{D},\theta)]-\mathbb{E}_{q}[\hat{w}(\theta)\nabla_{\lambda}\log q(\theta;\lambda)]
\end{eqnarray*}
}In another way, we assume the posterior $p(\theta|\mathcal{D})$
and the joint distribution $p(\mathcal{D},\theta)$ has no relation
to the parameter $\lambda$ of the variational distribution, that
is $\nabla_{\lambda}p(\theta|\mathcal{D})=0$ and $\nabla_{\lambda}\log p(\mathcal{D},\theta)=0$.
So we have,
\begin{eqnarray*}
\nabla_{\lambda}\mathcal{U} & = & \nabla_{\lambda}\int p(\theta|\mathcal{D})\log\frac{p(\mathcal{D},\theta)}{q(\theta;\lambda)}d\theta\\
 & = & \int\nabla_{\lambda}\left[p(\theta|\mathcal{D})(\log p(\mathcal{D},\theta)-\log q(\theta;\lambda))\right]d\theta\\
 & = & \int\nabla_{\lambda}p(\theta|\mathcal{D})(\log p(\mathcal{D},\theta)-\log q(\theta;\lambda))d\theta+\int p(\theta|\mathcal{D})\nabla_{\lambda}(\log p(\mathcal{D},\theta)-\log q(\theta;\lambda))d\theta\\
\\
 & = & -\int q(\theta;\lambda)\frac{p(\theta|\mathcal{D})}{q(\theta;\lambda)}\nabla_{\lambda}\log q(\theta;\lambda)d\theta\\
 & = & -\mathbb{E}_{q}[\hat{w}(\theta)\nabla_{\lambda}\log q(\theta;\lambda)]
\end{eqnarray*}
where $\hat{w}(\theta)=\frac{p(\theta|\mathcal{D})}{q(\theta;\lambda)}$,
in the last step we take importance sampling from $q(\theta;\lambda)$,
instead of directly sampling from $p(\theta|\mathcal{D})$.

\subsection*{The relation between KL-divergence and $\alpha$-divergence/$\chi^{n}$-divergence}

We show that the KL-divergence $D_{\mathrm{KL}}(p||q)$ is a tighter
divergence than $\alpha$-divergence $D_{\alpha}(q||p)$ (for $\alpha<0$).
By the Jensen's inequality, we known that for a concave function,
such as $f(\cdot)=\log(x)$ for $x>0$, we have $f(\mathbb{E}[x])\geq\mathbb{E}[f(x)]$.
So we have,
\begin{eqnarray*}
D_{\alpha}(q(\theta)||p(\theta|\mathcal{D})) & = & \frac{1}{1-\alpha}\log\int q(\theta)\left(\frac{p(\theta|\mathcal{D})}{q(\theta)}\right)^{1-\alpha}d\theta\\
 & = & \frac{1}{1-\alpha}\log\int p(\theta|\mathcal{D})\left(\frac{p(\theta|\mathcal{D})}{q(\theta)}\right)^{-\alpha}d\theta\\
 & \geq & \text{\ensuremath{\frac{1}{1-\alpha}\int p(\theta|\mathcal{D})\log}}\left[\left(\frac{p(\theta|\mathcal{D})}{q(\theta)}\right)^{-\alpha}\right]d\theta\\
 & = & \frac{-\alpha}{1-\alpha}D_{\mathrm{KL}}(p(\theta|\mathcal{D})||q(\theta)).
\end{eqnarray*}

Given this inequality, it is easy to find the relation between their
corresponding bounds,
\begin{align*}
\mathcal{U}_{\alpha}(\lambda) & =\frac{1}{1-\alpha}\log\mathbb{E}_{q}[(\frac{p(\mathcal{D},\theta)}{q(\theta;\lambda)})^{1-\alpha}]\\
 & =\frac{1}{1-\alpha}\log\mathbb{E}_{q}[(\frac{p(\theta|\mathcal{D})}{q(\theta;\lambda)})^{1-\alpha}]+\frac{1}{1-\alpha}\log p(\mathcal{D})^{1-\alpha}\\
 & \geq\frac{-\alpha}{1-\alpha}D_{\mathrm{KL}}(p(\theta|\mathcal{D})||q(\theta))+\log p(\mathcal{D})\\
 & =\frac{-\alpha}{1-\alpha}\int p(\theta|\mathcal{D})\log\dfrac{p(\mathcal{D},\theta)}{q(\theta;\lambda)}d\theta+\frac{1}{1-\alpha}\log p(\mathcal{D})\\
 & =\frac{-\alpha}{1-\alpha}\mathcal{U}(\lambda)+\frac{1}{1-\alpha}\log p(\mathcal{D}).
\end{align*}
Let $\alpha=1-n$, then we get similar inequalities between the $\chi^{n}$-divergence
$D_{\chi^{n}}(q||p)$ and the KL-divergence $D_{\mathrm{KL}}(p||q)$,
and their corresponding bounds.
\end{document}